\documentclass[]{youtu} % For LaTeX2e

\usepackage{mathpazo}
\usepackage{graphicx}
\usepackage{natbib}
\usepackage[caption=false]{subfig}
\usepackage{multirow}
\usepackage{colortbl}
\usepackage{wrapfig}
\usepackage{booktabs}
\usepackage{caption}
\usepackage[linesnumbered,ruled,vlined]{algorithm2e}
\usepackage{subcaption}
\usepackage{listings}
\newcommand\blfootnote[1]{%
  \begingroup
  \renewcommand\thefootnote{}\footnote{#1}%
  \addtocounter{footnote}{-1}%
  \endgroup
}

\title{HiChunk: Evaluating and Enhancing Retrieval-Augmented Generation with Hierarchical Chunking}
\author{
    Wensheng Lu \textsuperscript{*\ 1}
    \quad Keyu Chen \textsuperscript{*\ 1} 
    \quad Ruizhi Qiao \textsuperscript{1} 
    \quad Xing Sun \textsuperscript{1}
}
\affiliation{
    \textsuperscript{1}Tencent Youtu Lab\\[2pt]
}

\date{Sep 15, 2025}
\correspondence{Ruizhi.Qiao@tencent.com}
\sourcecode{https://github.com/TencentYoutuResearch/HiChunk.git}
\data{https://huggingface.co/datasets/Youtu-RAG/HiCBench}
\begin{document}
\blfootnote{$^{*}$ Equal Contribution}

\abstract{Retrieval-Augmented Generation (RAG) enhances the response capabilities of language models by integrating external knowledge sources. However, document chunking as an important part of RAG system often lacks effective evaluation tools. This paper first analyzes why existing RAG evaluation benchmarks are inadequate for assessing document chunking quality, specifically due to evidence sparsity. Based on this conclusion, we propose HiCBench, which includes manually annotated multi-level document chunking points, synthesized evidence-dense question answer(QA) pairs, and their corresponding evidence sources. Additionally, we introduce the HiChunk framework, a multi-level document structuring framework based on fine-tuned LLMs, combined with the Auto-Merge retrieval algorithm to improve retrieval quality. Experiments demonstrate that HiCBench effectively evaluates the impact of different chunking methods across the entire RAG pipeline. Moreover, HiChunk achieves better chunking quality within reasonable time consumption, thereby enhancing the overall performance of RAG systems.}
\maketitle

\vspace{-.1em}

\section{Introduction}
RAG (Retrieval-Augmented Generation) enhances the quality of LLM responses to questions beyond their training corpus by flexibly integrating external knowledge through the retrieval of relevant content chunks as prompts\citep{lewis2020retrieval}. This approach helps reduce hallucinations\citep{chen2024benchmarking, zhang2025siren}, especially when dealing with real-time information\citep{he2022rethinking} and specialized domain knowledge\citep{wang2023survey, li2023chatgpt}. Document chunking, a crucial component of RAG systems, significantly impacts the quality of retrieved knowledge and, consequently, the quality of responses. Poor chunking methods may separate continuous fragments, leading to information loss, or combine unrelated information, making it more challenging to retrieve relevant content. For instance, as noted in \cite{bhat2025rethinking}, the optimal chunk size varies significantly across different datasets.

Although numerous benchmarks exist for evaluating RAG systems\citep{bai-etal-2024-longbench, dasigi-etal-2021-dataset, duarte-etal-2024-lumberchunker, zhang2024ocr, yang2018hotpotqa, kovcisky2018narrativeqa, pang2021quality}, they mostly focus on assessing either the retriever's capability or the reasoning ability of the response model, without effectively evaluating chunking methods. We analyzed several datasets to determine the average word and sentence count of evidence. As shown in \autoref{tab:dataset_statistics}, existing benchmarks generally suffer from evidence sparsity, where only a few sentences in the document are relevant to the query. As illustrated in \autoref{fig:illustration}, this sparsity of evidence makes these datasets inadequate for evaluating the performance of chunking methods. In reality, user tasks might be evidence-dense, such as enumeration or summarization tasks, requiring chunking methods to accurately and completely segment semantically continuous fragments. Therefore, it is essential to effectively evaluate chunking methods.

To address this, we introduce \textbf{Hi}erarchical \textbf{C}hunking Benchmark(\textbf{HiCBench}), a benchmark for document QA designed to effectively evaluate the impact of chunking methods on different components of RAG systems, including the performance of document chunking, retrievers, and response models. HiCBench's original documents are sourced from OHRBench. We curated documents of appropriate length for the corpus and manually annotated chunking points at various hierarchical levels for evaluation purposes. These points are used to assess the chunker's performance and construct QA pairs, followed by using LLMs and the annotated document structure to create evidence-dense QA, and finally extracting relevant evidence sentences and filtering non-compliant samples using LLMs.

Additionally, existing document chunking methods only consider linear document structure\citep{duarte-etal-2024-lumberchunker, 10.1145/3626772.3657878, zhao-etal-2025-moc, wang2025document}, while user problems may involve fragments with different semantic granularity, and linear document structure makes it difficult to adaptively adjust during retrieval. Therefore, we propose the \textbf{Hi}erarchical \textbf{C}hunking framework(\textbf{HiChunk}), which employs fine-tuned LLMs for hierarchical document structuring and incorporates iterative reasoning to address the challenge of adapting to extremely long documents. For hierarchically structured documents, we introduce the Auto-Merge retrieval algorithm, which adaptively adjusts the granularity of retrieval chunks based on the query, thereby maximizing retrieval quality.

In this work, our main contributions are as follows:
\begin{itemize}
    \item We introduce HiCBench, a benchmark designed to assess the performance of chunker and the impact of chunking methods on retrievers and response models within RAG systems. HiCBench includes information on chunking points at different hierarchical levels of documents, as well as sources of evidence and factual answers related to evidence-dense QA, enabling better evaluation of chunking methods.
    \item We propose the HiChunk framework, a document hierarchical structuring framework that allows RAG systems to dynamically adjust the semantic granularity of retrieval chunks.
    \item We conduct comprehensive performance evaluations on several open-source datasets and HiCBench, analyzing the impact of different chunking methods across three dimensions: performance of chunker, retriever, and responser.
\end{itemize}

\begin{minipage}{0.45\textwidth}
    \centering
    \captionof{table}{Statistics of document QA benchmark.}
    \label{tab:dataset_statistics}
    \small
    \begin{tabular}{lcccc}
        \toprule
        \textbf{Dataset} & \textbf{Qasper} & \textbf{OHRBench} & \textbf{GutenQA} \\
        \hline
        \bf $\mathrm{Num}_{doc}$ & 416 & 1261 & 100 \\
        \bf $\mathrm{Sent}_{d}$ & 164 & 176 & 5,373 \\
        \bf $\mathrm{Word}_{d}$ & 4.2k & 5.4k & 146.5k \\
        \hline
        \bf $\mathrm{Num}_{qa}$ & 1,372 & 8,498 & 3,000 \\
        \bf $\mathrm{Word}_{q}$ & 8.9 & 20.6 & 16.0 \\
        \bf $\mathrm{Word}_{a}$ & 16.0 & 5.6 & 26.0 \\
        \hline
        \rowcolor{gray!20}
        \bf $\mathrm{Word}_{e}$ & 239.4 & 36.5 & 39.3 \\
        \rowcolor{gray!20}
        \bf $\mathrm{Sent}_{e}$ & 10.5 & 1.7 & 1.7 \\
        \hline
    \end{tabular}
\end{minipage}
\begin{minipage}{0.5\textwidth}
    \centering
    \includegraphics[width=0.99\textwidth]{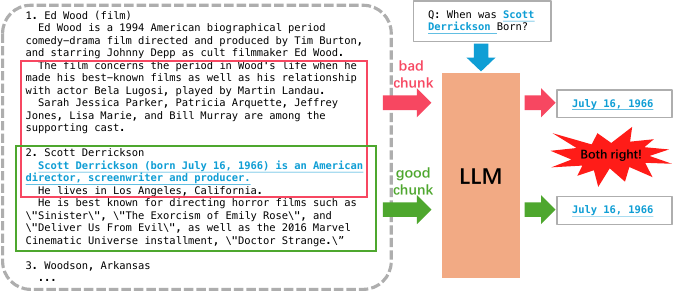}
    \captionof{figure}{Different chunk methods produce same answer.}
    \label{fig:illustration}
\end{minipage}

\section{Related Works}
\textbf{Traditional Text Chunking. }
Text chunking divides continuous text into meaningful units like sentences, phrases, and words, with our focus on sentence-level chunking.
Recent works have explored various approaches:
\citep{cho2022toward} combines text chunking with extractive summarization using hierarchical representations and determinantal point processes (DPPs) to minimize redundancy,
\citep{liu2021end} presents a pipeline integrating topical chunking with hierarchical summarization,
and \citep{zhang2021sequence} develops an adaptive sliding-window model for ASR transcripts using phonetic embeddings.
However, these LSTM and BERT\citep{devlin2019bert} based methods face limitations from small context windows and single-level chunking capabilities.

\textbf{RAG-oriented Document Chunking. } Recent research has explored content-aware document chunking strategies for RAG systems.
LumberChunker\citep{duarte-etal-2024-lumberchunker} uses LLMs to identify semantic shifts,but may miss hierarchical relationships. 
PIC\citep{wang2025document} proposes pseudo-instruction for document chunking, guide chunking via document summaries, 
though its single-level approach may oversimplify document structure. 
AutoChunker\citep{jain2025autochunker} employs tree-based representations but primarily focuses on noise reduction rather than multi-level granularity. 
Late Chunking\citep{gunther2024late} embeds entire documents before chunking to preserve global context, but produces flat chunk lists without modeling hierarchical relationships.
In contrast, our HierarchyChunk method creates multi-level document representations, 
chunking from coarse sections to fine-grained paragraphs.
This enables RAG systems to retrieve information at appropriate abstraction levels, 
effectively bridging fragmented knowledge gaps and providing comprehensive document understanding.

\textbf{Limitations of Existing Text Chunking Benchmarks. } 
The evaluation of text chunking and RAG methods heavily relies on benchmark datasets. 
Wiki-727\citep{koshorek-etal-2018-text},VT-SSum\citep{lv2021vt} and  NewsNet\citep{wu2023newsnet} 
are typically chunked into flat sequences of paragraphs or sentences, 
without capturing the multi-level organization (e.g., sections, subsections, paragraphs) inherent in many real-world documents.
This single-level representation limits the ability to evaluate chunking methods that aim to preserve or leverage document hierarchy, 
which is crucial for comprehensive knowledge retrieval in complex RAG scenarios.
While Qasper\citep{dasigi-etal-2021-dataset}, HotpotQA\citep{yang-etal-2018-hotpotqa} and GutenQA\citep{duarte-etal-2024-lumberchunker} are designed for RAG-related tasks, 
they do not specifically provide mechanisms or metrics for evaluating the efficacy of document chunking strategies themselves. 
Their focus is primarily on end-to-end RAG performance, where the impact of chunking is implicitly measured through retrieval and generation quality. 
This makes it challenging to isolate and assess the performance of different chunking methods independently, 
hindering systematic advancements in hierarchical document chunking. 
Our work addresses these gaps by proposing a method that explicitly considers multi-level document chunking
and constructs a novel benchmark from a chunking perspective.

\section{HiCBench Construction}
In order to construct the HiCBench dataset, we performed additional document hierarchical structuring and created QA pairs to evaluate document chunking quality, building on the OHRBench document corpus\citep{zhang2024ocr}. We filter documents with fewer than 4,000 words and those exceeding 50 pages. For retained documents, we manually annotated the hierarchical structure and used these annotations to assist in the generation of QA pairs and to assess the accuracy of document chunking.

\paragraph{Task Criteria} To ensure that the constructed QA pairs could effectively evaluate the quality of document chunking, we aimed for the evidence associated with each QA pair to be widely distributed across a complete semantic chunk. Failure to fully recall such a semantic chunk would result in missing evidence, thereby degrading the quality of the generated responses. To achieve this objective, we established the following standards to regulate the generation of QA pairs:
\begin{itemize}
    \item \textbf{Evidence Completeness and Density}: Evidence completeness ensures that the evidence relevant to the question is comprehensive and necessary within the context. Evidence density requires that evidence constitutes a significant proportion of the context, enhancing the QA pair's utility for evaluating chunking methods.
    \item \textbf{Fact Consistency}: To ensure the constructed samples can evaluate the entire retrieval-based pipeline, it is essential that the generated responses remain consistent with the answers when provided with full context, and that the questions are answerable.
\end{itemize}

\paragraph{Task Definition} Additionally, we define three different task types to evaluate the quality of chunking:
\begin{itemize}
    \item \textbf{Evidence-Sparse QA ($T_0$)}: The evidence related to the QA is confined to one or two sentences within the document.
    \item \textbf{Single-Chunk Evidence-Dense QA ($T_1$)}: Evidence sentences related to the QA constitute a substantial portion of the context within a single complete semantic chunk. The size of chunk between 512 and 4096 words.
    \item \textbf{Multi-Chunk Evidence-Dense QA ($T_2$)}: Evidence sentences related to the QA are distributed across multiple complete semantic chunks, covering a significant portion of the context. The size of chunk between 256 and 2048 words.
\end{itemize}

\paragraph{QA Construction} We use a prompt-based approach using DeepSeek-R1-0528 to generate candidate QA pairs, followed by a series of filtering processes to ensure the retained QA pairs meet the criteria of evidence completeness, density, and fact consistency. The specific process is as follows:
\begin{enumerate}
    \item \textbf{Document Hierarchical Annotation and Summarization}: To enable LLMs to gain an overall understanding of the specific document $D$ while constructing QA pairs, we first generated summaries for corresponding sections based on the annotated hierarchical structure, denoted as $S \gets LLM_{s}(D)$. These summaries will be used in QA pair generation.
    \item \textbf{Generation of Questions and Answers}: We randomly selected one or two chunks from all eligible document fragments as context \( C \), then generated candidate QA pairs using \( (S, C) \), where \( (Q, A) \leftarrow LLM_{qa}(S, C) \).
    \item \textbf{Ensuring Evidence Completeness and Density}: Referring to \cite{friel2024ragbench}, we use LLMs to extracted sentences from context \( C \) related to the QA pair as evidence, denoted as \( E \gets LLM_{ee}(C, Q, A) \). To mitigate hallucination effects, this step will be repeated five times, retaining sentences that appeared at least four times as the final evidence. Furthermore, to ensure evidence density, we remove samples which the ratio of evidence is less than 10\% of context \( C \).
    \item \textbf{Ensuring Fact Consistency}: We applied Fact-Cov metric\citep{xiang2025use} to filter test samples. We first extract the facts from answer $A$, denoted as $F \gets LLM_{fe}(Q, A)$\footnote[1]{\url{https://github.com/GraphRAG-Bench/GraphRAG-Benchmark}\label{myfootnote}}. Contexts $C$ used for constructing QA pairs will be provided to LLMs to generate response $R'$, denoted as $R' \gets LLM_{r}(Q, C)$. Then, the Fact-Cov metric will be calculated by $\mathrm{Fact\_Cov} \gets LLM_{fc}(F, R')$\textsuperscript{\ref{myfootnote}}. This process will be repeated 5 times. We retain samples with an average Fact-Cov metric exceeding 80\%. Samples below this threshold are deemed unanswerable. All prompts used for QA construction are provided in \autoref{app:prompts}.
\end{enumerate}

\section{Methodology}

\begin{figure}[t]
    \centering
    \includegraphics[width=0.99\textwidth]{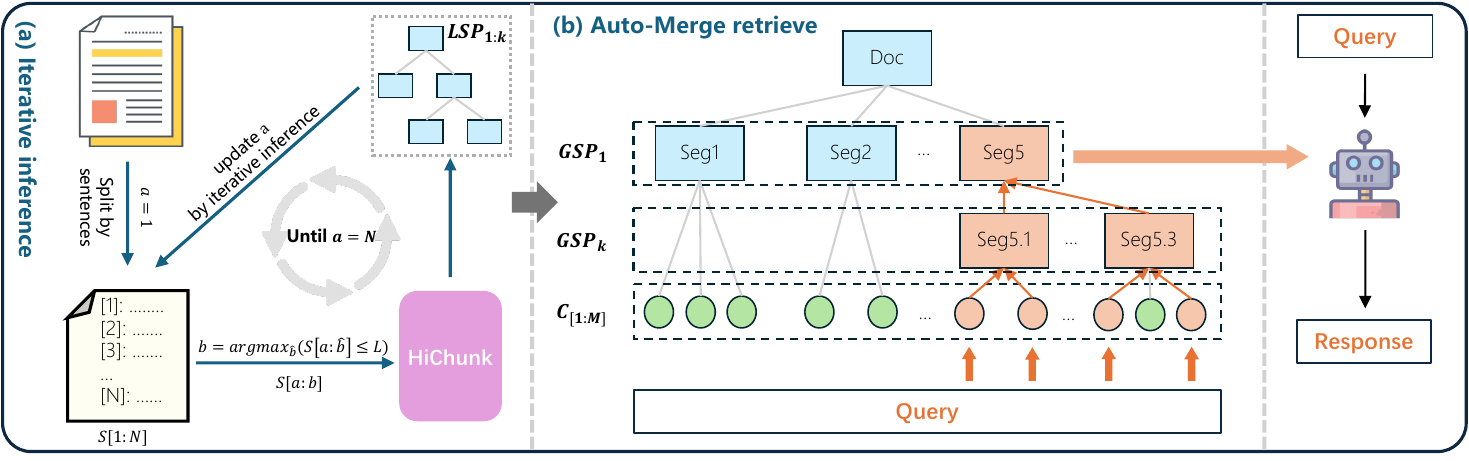}
    \captionof{figure}{Framework. (a) Iterative inference for HiChunk on long documents. (b) Auto-Merge retrieval algorithm.}
    \label{fig:framework}
\end{figure}

This section primarily introduces the HiChunk framework. The overall framework is illustrated in \autoref{fig:framework}. The aim is for the fine-tuned LLMs to comprehend the hierarchical relationships within a document and ultimately organize the document into a hierarchical structure. This involves two subtasks: identification of chunking points and determination of hierarchy levels. Through prompt, HiChunk converts these two subtasks into text generation task. In model train of HiChunk, we use Gov-report\citep{huang-etal-2021-efficient}, Qasper\citep{dasigi-etal-2021-dataset} and Wiki-727\citep{koshorek-etal-2018-text} to construct training instructions, which are publicly available datasets with explicit document structure. Meanwhile, we augment the training set by randomly shuffling document chapters and deleting document content.
During inference, HiChunk begins by processing document \(D\) with sentence chunking, assigning each sentence an independent ID, resulting in \(S[1:N]\), where \(N\) is the number of sentences. Through HiChunk, the document's global chunk points \(GCP_{1:k}\) are obtained, denoted as \(GCP_{1:k} \gets \mathrm{HiChunk}(S[1:N])\), with \(k\) being the maximum number of predicted chunk point levels, and \(GCP_i\) representing all chunk points at level \(i\). Additionally, an iterative inference optimization strategy is proposed to handle the structuring of extremely long documents.

Although the HiChunk hierarchical tree structure has semantic integrity, the variability in the chunk length distribution caused by the semantic chunking method can lead to disparities in semantic granularity, which can affect retrieval quality. To mitigate this, we apply a fixed-size chunking approach on the results of HiChunk to produce $C[1:M]$, and propose the Auto-Merge retrieval algorithm to balance issues of varying semantic granularity and the semantic integrity of retrieved chunks.

\paragraph{Iterative Inference}
For documents that exceed the predefined inference input length \(L\), an iterative inference approach is required. The process begins by initializing the start sentence id of the current iteration as \(a=0\) and determining the end sentence id \(b\) to ensure the constructed input length is less than the maximum predefined length \(L\). This is optimally set as \(b = \arg\max_{\hat{b}}(S[a:\hat{b}] < L)\). Through this setup, local chunk points \(LCP_{1:k} \gets \mathrm{HiChunk}(S[a:b])\) are obtained via inference. These local chunk points \(LCP_{1:k}\) are then merged into the global chunk points \(GCP_{1:k}\). Based on these local inference results, the new value of \(a\) is determined for the next iteration. This iterative process continues until \(a = N\), signifying the completion of inference for the entire document. 
However, iterative inference suffers from hierarchical drift when there is only one level 1 chunk in the local inference result. To mitigate this problem, we construct residual text lines from known document structures to guide the model making correct hierarchical judgments.
The complete iterative inference procedure is illustrated in \autoref{alg:iterative_inference}.

\paragraph{Auto-Merge Retrieval Algorithm}

To balance the semantic richness and completeness of recalled contexts, we propose Auto-Merge retrieval algorithm. This algorithm uses a series of conditions to control the extent to which child nodes are merged upward into parent nodes. Auto-Merge algorithm traverses the query-ranked chunks \( C_{[1:M]}^{sorted} \), using \(\mathrm{node_{ret}}\) to record the nodes that have been recalled. During the \(i\)-th step of the traversal, we first record the current used token budget, \( tk_{cur} = \sum_{n \in \mathrm{node_{ret}}}^{} \mathrm{len}(n) \). We then add \( C_{[i]}^{sorted} \) to \(\mathrm{node_{ret}}\) and denote the parent of \( C_{[i]}^{sorted} \) by \(p\). Finally, we merge upward when the following conditions are met:
\begin{itemize}
    \item $Cond_1$: The intersection between the children of \( p \) and \(\mathrm{node_{ret}}\) contains at least two elements, denoted as \(\sum_{n \in \mathrm{node_{ret}}}^{} \mathbb{1}(n \in p.\mathrm{children}) \geq 2\).
    \item $Cond_2$: The text length of the nodes in \(\mathrm{node_{ret}}\) belonging to \( p \)'s children is greater than or equal to \(\theta^*\), denoted as \(\sum_{n \in (\mathrm{node_{ret}} \ \cap \ p.\mathrm{children})} \mathrm{len}(n) \geq \theta^*\). Here, \(\theta^*\) is an adaptively adjusted threshold, defined as \(\theta^*(tk_{cur}, p) = \frac{\mathrm{len}(p)}{3} \times (1+\frac{tk_{cur}}{T})\). As \( tk_{cur} \) increases, \(\theta^*\) gradually increases from \(\frac{\mathrm{len}(p)}{3}\) to \(\frac{2 \times \mathrm{len}(p)}{3}\). This means that higher-ranking chunks are more likely to merge upward.
    \item $Cond_3$: The remaining token budget is greater than the length of the chunk corresponding to \( p \), denoted as \( T - tk_{cur} \geq \mathrm{len}(p) \).
\end{itemize}
The entire retrieval algorithm process is illustrated in \autoref{alg:retrieval_algorithm}.

\begin{minipage}{0.46\textwidth}
    \begin{algorithm}[H]
    \caption{iterative inference}
    \label{alg:iterative_inference} 
    \SetKw{KwReturn}{return}
    \SetKw{KwBreak}{break}
    \SetKw{KwAnd}{and}
    \SetKwData{a}{$a$}
    \SetKwData{b}{$b$}
    \SetKwData{S}{$S[1:N]$}
    \SetKwData{ResidualLines}{$res\_lines$}
    \SetKwData{LocalChunkPoint}{$LCP_{1:k}$}
    \SetKwData{GlobalChunkPoint}{$GCP_{1:k}$}
    \SetKwFunction{Merge}{Merge}
    \SetKwFunction{HiChunk}{HiChunk} 
    \SetKwFunction{SentenceTokenize}{SentTokenize} 
    \SetKwFunction{GetResLines}{GetResLines} 
    \SetKwInOut{Input}{input}
    \SetKwInOut{Output}{output}
	
    \Input{Document $D$, Inference lengths $L$}
    \Output{Global chunk points \GlobalChunkPoint}
    \BlankLine
    \S $\leftarrow$ \SentenceTokenize{$D$}\;
    \a $\leftarrow$ 1\;
    \b $\leftarrow \mathrm{argmax}_{\hat{b}}(S[a:\hat{b}] \leq L)$\;
    \ResidualLines $\leftarrow$ None\;
    \GlobalChunkPoint $\leftarrow []*k$ \;
    \While{$1 \leq a < b \leq N$}{
       \LocalChunkPoint $\leftarrow$ \HiChunk{$S[a:b]$, \ResidualLines}\;
        % \tcp{merge \LocalChunkPoint into \GlobalChunkPoint}
       \GlobalChunkPoint $\leftarrow$ \Merge{\GlobalChunkPoint, \LocalChunkPoint}\;
        \If{$len(LCP_1) \geq 2$}{
            \tcp{the last chunk point at first level}
            \a $\leftarrow LCP_1[-1]$\;
            \ResidualLines $\leftarrow$ None\;
        }
        \Else{
            \a $\leftarrow$ \b\;
            \ResidualLines $\leftarrow$ GetResLines(\GlobalChunkPoint)\;
        }
        \b $\leftarrow \mathrm{argmax}_{\hat{b}}(S[a:\hat{b}] \leq L)$\;
    }
    \KwReturn \GlobalChunkPoint
    \end{algorithm}
\end{minipage}\hfill
\begin{minipage}{0.52\textwidth}
    \begin{algorithm}[H]
    \caption{retrieval algorithm}
    \label{alg:retrieval_algorithm} 
    \SetKw{KwReturn}{return}
    \SetKw{KwBreak}{break}
    \SetKw{KwAnd}{and}
    \SetKwData{P}{$p$}
    \SetKwData{Node}{$\mathrm{node_{ret}}$}
    \SetKwData{TkCur}{$tk_{cur}$}
    \SetKwData{Context}{ctx}
    \SetKwFunction{Merge}{Merge}
    \SetKwFunction{FindParent}{FindParent} 
    \SetKwFunction{BuildContext}{BuildContext} 
    \SetKwFunction{AdaptiveTheta}{$\theta^*$} 
    \SetKwFunction{Sorted}{Sorted} 
    \SetKwInOut{Input}{input}
    \SetKwInOut{Output}{output}
    \Input{Token budget $T$, Chunks $C_{[1:M]}$, Query $q$}
    \Output{Retrieval context \Context}
    \BlankLine 
    $C^{sorted}_{[1:M]} \leftarrow$ \Sorted{$C_{[1:M]}, q$}\;
    \Node $\leftarrow$ $[]$, \TkCur $\leftarrow$ $0$\;
    % \emph{special treatment of the first line}\; 
    \For{$i\leftarrow 1$ \KwTo $M$}{
        \Node $\leftarrow$ \Node + $C_{[i]}^{sorted}$\;
        \Context, \TkCur $\leftarrow$ \BuildContext{\Node}\;
        \P $\leftarrow$ $C_{[i]}^{sorted}.parent$\; 
        \While{$Cond_1$ \KwAnd $Cond_2$ \KwAnd $Cond_3$}{
            \If{\TkCur $\geq T$}{
                \KwBreak
            }
            \tcp{add \P to \Node and remove nodes covered by \P}
            \Node $\leftarrow$ \Merge{\Node, \P}\;
            \Context, \TkCur $\leftarrow$ \BuildContext{\Node}\;
            \P $\leftarrow$ \P$.parent$\; 
        }
        \If{\TkCur $\geq T$}{
            \KwBreak
        }
    }
    \KwReturn \Context$[:T]$
    \end{algorithm}
\end{minipage}

\section{Experiments}
\subsection{Datasets and Metrics}
The test subsets of Gov-report\citep{huang-etal-2021-efficient} and Qasper\citep{dasigi-etal-2021-dataset} datasets will be used for evaluation of chunking accuracy. For the Gov-report dataset, we only retain documents with document word count greater than 5k for experiments. 
To evaluate the accuracy of the chunking points, we use the $F1$ metrics of the chunking points. The $F1_{L_1}$ and $F1_{L_2}$ correspond to the chunking points of the level 1 and level 2 chunks, respectively. And the $F1_{L_{all}}$ metric does not consider the level of the chunking point. The Qasper, GutenQA\citep{duarte-etal-2024-lumberchunker}, and OHRBench\citep{zhang2024ocr} datasets contain evidence relevant to the question. These datasets will be used in the evaluation for context retrieval. 

For the full RAG pipeline evaluation, we used the publicly available datasets LongBench\citep{bai-etal-2024-longbench}, Qasper, GutenQA, and OHRBench. the LongBench RAG evaluation contains 8 subsets from different datasets, with a total of 1,550 qa pairs, which can be categorized into single document qa and multiple document qa. The Qasper dataset contains 1,372 qa pairs from 416 documents. The GutenQA dataset contains 3,000 qa pairs based on 100 documents. In GutenQA, the average number of words in a document is 146,506, which is significantly higher than the other datasets. The documents of OHRBench come from seven different areas. We keep the documents with word counts greater than 4k in OHRBench and use the original qa pairs corresponding to these documents as a representative of the task $T_0$, denoted as OHRBench($T_0$). We use the F1 score and Rouge metrics to assess the quality of LLM responses. All experiments are conducted in the code repository of LongBench\footnote{\url{https://github.com/THUDM/LongBench/tree/main}}.

Furthermore, HiCBench will be used for comprehensive evaluation, including chunking accuracy, evidence recall rate, and RAG response quality assessment. To avoid biases from sparse text quality evaluation metrics, we employ the Fact-Cov\citep{xiang2025use} metric for response quality evaluation of HiCBench. The Fact-Cov metric is repeatedly calculated 5 times to take the average. Statistics information of datasets used in experiment are shown in \autoref{tab:dataset_statistics_exp}.
\begin{table}[htbp]
    \centering
    \captionof{table}{Statistics of dataset used in experiments.}
    \label{tab:dataset_statistics_exp}
    \small
    \begin{tabular}{lcccccc}
        \toprule
        \textbf{Dataset} & \textbf{Qasper}  &\textbf{GutenQA} & \textbf{OHRBench($T_0$)} & \textbf{HiCBench($T_1$, $T_2$)} \\
        \hline
        \bf $\mathrm{Num}_{doc}$ & 416 & 100 & 214 & 130 \\
        \bf $\mathrm{Sent}_{d}$ & 164 & 5,373 & 886 & 298 \\
        \bf $\mathrm{Word}_{d}$ & 4.2k & 146.5k & 26.8k & 8.5k \\
        \hline
        \bf $\mathrm{Num}_{qa}$ & 1,372 & 3,000 & 4,702 & (659, 541) \\
        \bf $\mathrm{Word}_{q}$ & 8.9 & 16.0 & 22.2 & (31.0, 33.0) \\
        \bf $\mathrm{Word}_{a}$ & 16.0 & 26.0 & 4.8 & (130.1, 126.4) \\
        \hline
        \rowcolor{gray!20}
        \bf $\mathrm{Word}_{e}$ & 239.4 & 39.3 & 39.1 & (561.5, 560.5) \\
        \rowcolor{gray!20}
        \bf $\mathrm{Sent}_{e}$ & 10.5 & 1.7 & 1.7 & (20.5, 20.4) \\
        \hline
    \end{tabular}
\end{table}

\subsection{Comparison Methods}
We primarily compared tow types of chunking methods: rule-based chunking methods and semantic-based chunking methods. All the comparison methods are as follows:
\begin{itemize}
    \item \textbf{FC200}: Fixed chunking is a rule-based method, which first divide the document into sentences and then merge sentences based on a fixed chunking size. Here, the fixed chunking size is 200.
    \item \textbf{SC}: Semantic chunker\citep{10.1145/3626772.3657878} uses an embedding model to calculate the similarity between adjacent paragraphs for chunking. We use bge-large-en-v1.5\citep{10.1145/3626772.3657878} as the embedding model.
    \item \textbf{LC}: LumberChunker\citep{duarte-etal-2024-lumberchunker} employs LLMs to predict the positions for chunking. In our experiments, we use Deepseek-r1-0528\citep{deepseekai2025deepseekr1incentivizingreasoningcapability} as the prediction model. The sampling temperature set to 0.1.
    \item \textbf{HC200}: Hierarchical chunker is the proposed method. In the model training for HiChunk. We further chunk the chunks of HiChunk by the fixed chunking method. The fixed chunking size is set to 200, denoted as HC200.
    \item \textbf{HC200+AM}: "+AM" represents the result of introducing Auto-Merge retrieval algorithm on the basis of HC200.
\end{itemize}

\subsection{Experimental Settings}
In the model training of HiChunk, Gov-report\citep{huang-etal-2021-efficient}, Qasper\citep{dasigi-etal-2021-dataset} and Wiki-727\citep{koshorek-etal-2018-text} are the train datasets, which are publicly available datasets with explicit document structure. We use Qwen3-4B\citep{qwen3technicalreport} as the base model, with a learning rate of 1e-5 and a batch size of 64. The maximum length of training and inference is set to 8192 and 16384 tokens, respectively. Meanwhile, the length of each sentence is limited to within 100 characters.
Due to the varying sizes of chunks resulting from semantic-based chunking, we limit the length of the retrieved context based on the number of tokens rather than the number of chunks for a fair comparison. The maximum length of the retrieved context is set to 4096 tokens. We also compare the performance of different chunking methods under different retrieved context length settings in \autoref{sec:retrieval_size_exp}. In the RAG evaluation process, we consistently use Bge-m3\citep{chen-etal-2024-m3} as the embedding model for context retrieval. As for the response model, we use three different series of LLMs with varying scales: Llama3.1-8b\citep{dubey2024llama}, Qwen3-8b, and Qwen3-32b\citep{qwen3technicalreport}.

\subsection{Chunking Accuracy}
To comprehensively evaluate the performance of the semantic-based chunking method, we conducted experiments using two publicly available datasets, along with the proposed benchmark, to assess the cut-point accuracy of the chunking method. Since the SC and LC chunking methods are limited to performing single-level chunking, we evaluated only the F1 scores for the initial level of chunking points and the F1 scores without regard for the hierarchy of chunking points. The evaluation results are presented in \autoref{tab:chunk_accuracy}. In the Qasper and Gov-report datasets, which serve as in-domain test sets, the HC method shows a significant improvement in chunk accuracy compared to the SC and LC methods. Additionally, in HiCBench, an out-of-domain test set, the HC method exhibits even more substantial accuracy improvements. These findings demonstrate that HC enhances the base model's performance in document chunking by focusing exclusively on the chunking task. Moreover, as indicated in the subsequent experimental results presented in \autoref{sec:rag_pipeline_eval}, the accuracy improvement of the HC method in document chunking leads to enhanced performance throughout the RAG pipeline. This includes improvements in the quality of evidence retrieval and model responses.

\begin{table}[htbp]
    \caption{Chunking accuracy. $\mathbf{HC}$ means the result of HiChunk without fixed size chunking. The best result is in \textbf{bold}.}
    \centering
    \resizebox{0.8\textwidth}{!}{
        \begin{tabular}{l|ccc|ccc|ccc}
            \toprule
            \multicolumn{1}{l|}{\bf Chunk}  &\multicolumn{3}{c|}{\bf Qasper} &\multicolumn{3}{c|}{\bf Gov-Report} &\multicolumn{3}{c}{\bf HiCBench} \\
            \multicolumn{1}{l|}{\bf Method} &$F1_{L_1}$  &$F1_{L_2}$ &$F1_{L_{all}}$ &$F1_{L_1}$  &$F1_{L_2}$ &$F1_{L_{all}}$ &$F1_{L_1}$  &$F1_{L_2}$ &$F1_{L_{all}}$ \\
            \midrule
            % Fix-Size Chunk      &- &-  &- &- &- &- &- &- &- \\
            SC      &0.0759 &-  &0.1007 &0.0298 &- &0.0616 &0.0487 &- &0.1507 \\
            LC      &0.5481 &-  &0.6657 &0.1795 &- & 0.5631 &0.2849 &- &0.4858 \\
            HC      &\bf 0.6742 &\bf 0.5169 &\bf 0.9441 &\bf0.9505 &\bf0.8895 &\bf0.9882 &\bf 0.4841 &\bf 0.3140 &\bf 0.5450 \\
            \bottomrule
        \end{tabular}
    }
    \label{tab:chunk_accuracy}
\end{table}

\subsection{RAG-pipeline Evaluation}
\label{sec:rag_pipeline_eval}

We evaluated the performance of various chunking methods on the LongBench, Qasper, GutenQA, OHRBench and HiCBench datasets, with the results detailed in \autoref{tab:bench_rag_eval}. The performance of each subset in LongBench is shown in \autoref{tab:LongBench_rag_eval}. The results demonstrate that the HC200+AM method achieves either optimal or suboptimal performance on most LongBench subsets. When considering average scores, LumberChunk remains a strong baseline. 
However, as noted in \autoref{tab:dataset_statistics_exp}, both GutenQA and OHRBench datasets exhibit the feature of evidence sparsity, meaning that the evidence related to QA pairs is derived from only a few sentences within the document. Consequently, the different chunking methods show minimal variation in evidence recall and response quality metrics on these datasets. For instance, using Qwen3-32B as the response model on the GutenQA dataset, the evidence recall metrics of FC200 and HC200+AM are 64.5 and 65.53, and the Rouge metrics are 44.86 and 44.94, respectively. Another example is OHRBench dataset, the evidence recall metrics and Rouge metrics of FC200, LC, HC200 and HC200+AM are very close.
In contrast, the Qasper and HiCBench datasets contain denser evidence, where a better chunking method results in higher evidence recall and improved response quality. Again using Qwen3-32B as an example, on the $T_1$ task of HiCBench dataset, the evidence recall metric for FC200 and HC200+AM are 74.06 and 81.03, the Fact-Cov metrics are 63.20 and 68.12, and the Rouge metrics are 35.70 and 37.29, respectively.
These findings suggest that the evidence-dense QA in the HiCBench dataset is better suited for evaluating the quality of chunking methods, enabling researchers to more effectively identify bottlenecks within the overall RAG pipeline.

\begin{table}[htbp]
    \caption{RAG-pipeline evaluation on LongBench, Qasper, GutenQA, OHRBench and HiCBench. Evidence Recall and Fact Coverage metric is represented by ERec and FC respectively. The best result is in \textbf{bold}, and the sub-optimal result is in \underline{underlined}}
    \centering
    \resizebox{\textwidth}{!}{
        \begin{tabular}{l|c|cc|cc|cc|ccc|ccc}
            \toprule
            \multicolumn{1}{l|}{\bf Chunk}  &\multicolumn{1}{c|}{\bf LongBench}  &\multicolumn{2}{c|}{\bf Qasper} &\multicolumn{2}{c|}{\bf GutenQA} &\multicolumn{2}{c|}{\bf OHRBench($T_0$)} &\multicolumn{3}{c|}{\bf HiCBench($T_1$)} &\multicolumn{3}{c}{\bf HiCBench($T_2$)} \\
            \multicolumn{1}{l|}{\bf Method}  &Score  &ERec  &F1 &ERec  &Rouge  &ERec  &Rouge  &ERec  &FC  &Rouge  &ERec  &FC  &Rouge \\
            \midrule
            \multicolumn{14}{c}{\bf Llama3.1-8B}  \\
            \hline
            FC200  &42.49  &84.08  &47.26  &64.43  &30.03  &67.03  &51.01  &74.84  &47.82  &28.43  &74.61  &46.79  &30.97    \\
            SC  &42.12     &82.08  &47.47  &58.30  &28.58  &62.65  &49.10  &72.14  &46.80  &28.43  &73.49  &45.28  &30.92      \\
            LC  &42.73     &\underline{87.08}  &\underline{48.20}  &63.67  &\underline{30.22}  &\bf68.42  &\underline{51.85}  &76.64  &\underline{50.84}  &\underline{29.62}  &76.12  &\underline{49.12}  &\underline{32.01}      \\
            \rowcolor{gray!30}
            HC200  &\bf43.17  &86.16  &48.09  &\underline{65.13}  &29.95  &\underline{68.25}  &51.33  &\underline{78.52}  &49.87  &29.38  &\underline{78.76}  &49.11  &31.80      \\
            \rowcolor{gray!30}
            \hspace{0.2cm}+AM  &\underline{42.90}      &\bf87.49  &\bf48.95  &\bf65.47   &\bf30.33  &67.84  &\bf51.92  &\bf81.59  &\bf55.58  &\bf30.04  &\bf80.96  &\bf53.66  &\bf33.04     \\
            \toprule
            \multicolumn{14}{c}{\bf Qwen3-8B}  \\
            \hline
            FC200  &43.95   &84.32          &45.10      &64.50          &33.47          &67.07      &48.18  &74.06  &47.35  &33.83  &72.95  &43.45  &35.27      \\
            SC  &43.54      &82.22          &44.55      &58.37          &32.71          &62.18      &46.79  &71.42  &46.07  &33.30  &72.36  &42.97  &34.76     \\
            LC  &\bf44.83      &\underline{87.43}  &\bf46.05  &63.67  &33.87  &\bf68.79  &\underline{49.28}  &75.53  &\underline{48.27}  &34.12  &75.14  &\underline{46.80}  &35.93    \\
            \rowcolor{gray!30}
            HC200  &43.90   &86.49          &\underline{45.95}      &\underline{65.20}  &\underline{33.89}  &\underline{68.57}      &49.06  &\underline{77.68}  &47.37  &\underline{34.30}  &\underline{78.10}  &46.20  &\underline{36.32}      \\
            \rowcolor{gray!30}
            \hspace{0.2cm}+AM  &\underline{44.41}     &\bf87.85       &45.82      &\bf65.53  &\bf34.15    &68.31      &\bf49.61  &\bf81.03  &\bf50.75  &\bf35.26  &\bf80.65  &\bf49.02  &\bf37.28      \\
            \toprule
            \multicolumn{14}{c}{\bf Qwen3-32B}  \\
            \hline
            FC200  &46.33   &84.32      &46.49      &64.50   &\underline{44.86}      &67.07      &46.89  &74.06  &63.20  &35.70  &72.95  &60.87  &37.17      \\
            SC  &46.29      &82.22   &46.39      &58.37      &43.59      &62.18  &45.43  &71.26  &61.09  &35.64  &72.36  &59.23  &37.09 \\
            LC  &\bf47.43      &\underline{87.43}      &46.82      &63.67      &44.45  &\bf68.79     &\bf47.92  &75.53  &\underline{64.76}  &36.15  &75.14  &\underline{62.75}  &38.02      \\
            \rowcolor{gray!30}
            HC200  &46.71   &86.49      &\underline{46.99}      &\underline{65.20}      &44.83      &\underline{68.57}      &47.71  &\underline{77.68}  &63.93  &\underline{36.55}  &\underline{78.10}  &62.51  &\underline{38.26}     \\
            \rowcolor{gray!30}
            \hspace{0.2cm}+AM  &\underline{46.92}     &\bf87.85      &\bf47.25   &\bf65.53      &\bf44.94  &68.31  &\underline{47.89}  &\bf81.03  &\bf68.12  &\bf37.29  &\bf80.65  &\bf66.36  &\bf37.37     \\
            \hline
        \end{tabular}
    }
    \label{tab:bench_rag_eval}
\end{table}

\subsection{Influence of Retrieval Token Budget}
\label{sec:retrieval_size_exp}
\begin{figure}[htbp]
    \centering
    \includegraphics[width=0.99\textwidth]{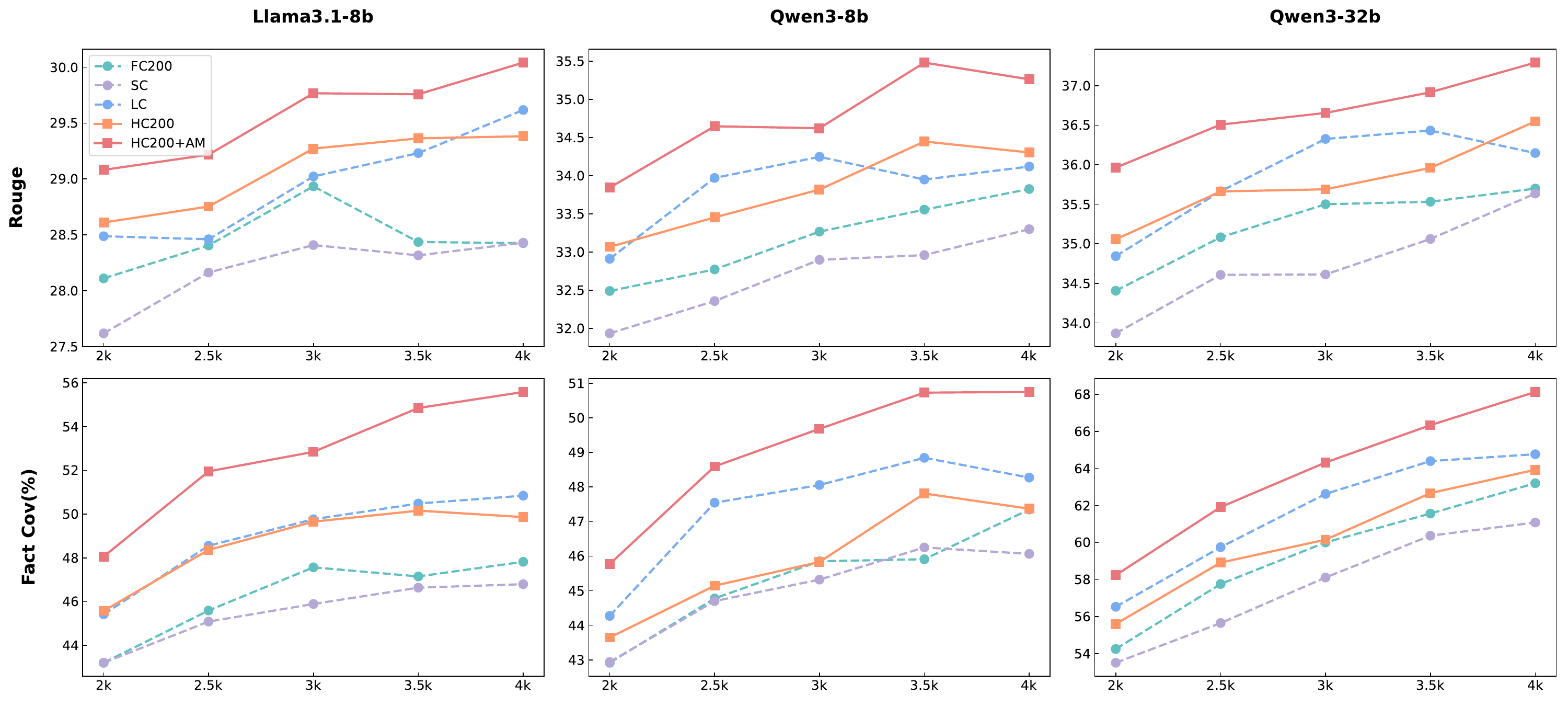}
    \caption{Performance of HiCBench($T_1$) under different retrieval token budget from 2k to 4k.}
    \label{fig:context_size_exp}
\end{figure}

Since HiCBench is more effective in assessing the performance of chunking methods, we evaluated the impact of our proposed method on the $T_1$  task of HiCBench under different retrieve token budgets: 2k, 2.5k, 3k, 3.5k and 4k tokens. We compared the effects of various chunking methods by calculating the Rouge metrics between responses and answers, as well as the Fact-Cov metrics. The experimental findings are illustrated in \autoref{fig:context_size_exp}. The results demonstrate that a larger retrieval token budget usually leads to better response quality, so it is necessary to compare different chunking methods under the same retrieval token budget. HC200+AM consistently achieves superior response quality across various retrieve token budget settings. These experimental results underscore the effectiveness of HC200+AM method. We further present the correspond curves of the evidence recall metrics in \autoref{sec:context_size_er_exp}.

\subsection{Effect of Maximum Hierarchical Level}
\begin{figure}[htbp]
    \centering
    \includegraphics[width=0.99\textwidth]{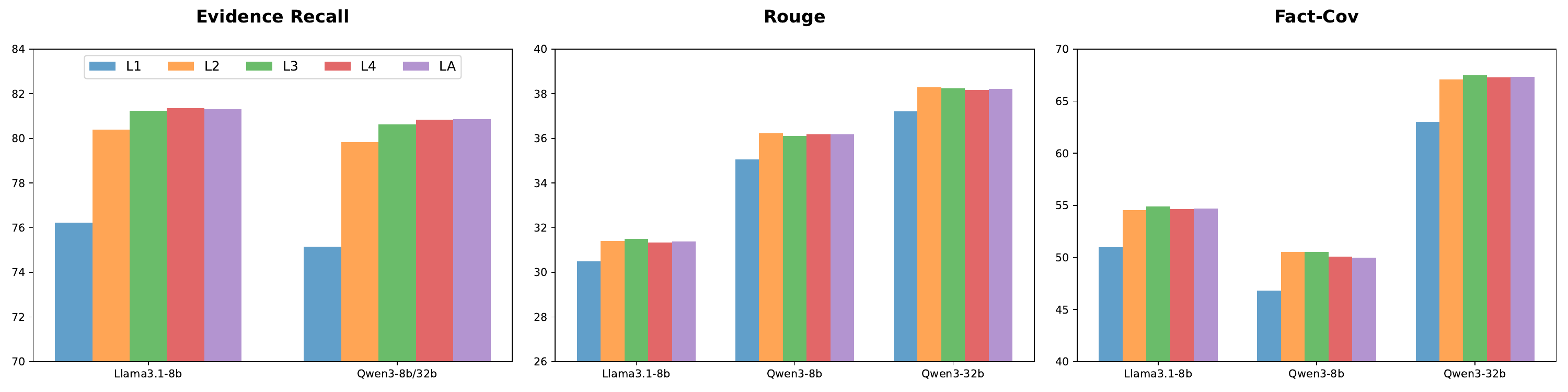}
    \caption{Evidence recall metric across different maximum level on  HiCBench($T_1$ and $T_2$).}
    \label{fig:level_study_exp}
\end{figure}

In this section, we examine the impact of limiting the maximum hierarchical level of document structure obtained by HiChunk. The maximum level ranges from 1 to 4, denoted as \(L1\) to \(L4\), while \(LA\) represents no limitation on the maximum level. We measure the evidence recall metric on different settings. As shown in \autoref{fig:level_study_exp}. This result reveals that the Auto-Merge retrieval algorithm degrades the performance of RAG system in the \(L1\) setting due to the overly coarse-grained semantics of \(L1\) chunks. As the maximum level increases from 1 to 3, the evidence recall metric also gradually improves and remains largely unchanged thereafter. These findings highlight the importance of document hierarchical structure for enchancing RAG systems.

\subsection{Time Cost for Chunking}
As document chunking is essential for RAG systems, it must meet specific timeliness requirements. In this section, we analyze the time costs associated with different semantic-based chunking methods, as presented in \autoref{tab:time_cost}. Although the SC method exhibits superior real-time performance, it consistently falls short in quality across various datasets compared to other baselines. However, the LC method demonstrates reasonably good performance, but its chunking speed is considerably slower than other semantic-based methods, limiting its applicability within RAG systems. In contrast, the HC method achieves the highest chunking quality among all baseline methods while maintaining an acceptable time cost, making it well suited for implementation in real scenarios RAG systems.

\begin{table}[htbp]
    \caption{Time cost of different chunking methods.}
    \centering
    \resizebox{0.9\textwidth}{!}{
        \begin{tabular}{l|c|cc|cc|cc}
            \toprule
            \multirow{2}*{\bf Dataset}  &\multirow{2}*{\bf Avg. Word} &\multicolumn{2}{c|}{\bf SC} &\multicolumn{2}{c|}{\bf LC} &\multicolumn{2}{c}{\bf HC}  \\
            & &Time(s/doc) &Chunks &Time(s/doc) &Chunks &Time(s/doc) &Chunks \\
            \midrule
            Qasper      &4,166   &0.4867  &43.83   &5.4991  &18.32  &1.4993  &15.08   \\
            Gov-report  &13,153  &1.3219  &114.72  &15.4321 &40.89  &4.3382   &29.79   \\
            OHRBench($T_0$)    &26,808  &3.0943  &249.14  &37.3935  &89.68  &14.5776  &92.23   \\
            GutenQA     &146,507 &16.5028 &1,453.00    &132.4900  &393.52  &60.1921  &232.85   \\
            HiCBench    &8,519 &1.0169  &80.12    &13.4414  &41.48  &5.7506  &51.35   \\
            \bottomrule
        \end{tabular}
    }
    \label{tab:time_cost}
\end{table}

\section{Conclusion}
This paper begins by analyzing the shortcomings of current benchmarks used for evaluating RAG systems, specifically highlighting how evidence sparsity makes them unsuitable for assessing different chunking methods. As a solution, we introduce HiCBench, a QA benchmark focused on hierarchical document chunking, which effectively evaluates the impact of various chunking methods on the entire RAG process. Additionally, we propose the HiChunk framework, which, when combined with the Auto-Merge retrieval algorithm, significantly enhances the quality of chunking, retrieval, and model responses compared to other baselines.

\setcitestyle{numbers,square}
\setcitestyle{square,numbers,comma}
\bibliography{youtu}

\clearpage
\appendix
\section{Appendix}
\definecolor{backcolour}{rgb}{0.96, 0.96, 0.96}
\definecolor{codegreen}{rgb}{0,0.6,0}
\lstdefinestyle{myStyle}{
    backgroundcolor=\color{backcolour},   
    commentstyle=\color{codegreen},
    basicstyle=\ttfamily\footnotesize,
    breakatwhitespace=true,         
    breaklines=true,                 
    keepspaces=true,                 
    numbers=none,       
    numbersep=5pt,                  
    showspaces=false,                
    showstringspaces=false,
    showtabs=false,                  
    tabsize=2,
    frame=single,
}
\lstset{style=myStyle}

\setcounter{table}{0}
\setcounter{figure}{0}
\setcounter{equation}{0}
\renewcommand{\thetable}{A\arabic{table}}
\renewcommand{\thefigure}{A\arabic{figure}}
\renewcommand{\theequation}{A\arabic{equation}}
\renewcommand{\thelstlisting}{A\arabic{lstlisting}}

\subsection{Detail of LongBench}
In this section, we present the metric of each subset of the different chunking methods on LongBench, and the results are shown in \autoref{tab:LongBench_rag_eval}.

\begin{table}[htbp]
    \caption{RAG-pipeline evaluation on LongBench and each subset. The best result is in \textbf{bold}, and the sub-optimal result is in \underline{underlined}. Qasper* is the subset of LongBench.}
    \centering
    \resizebox{0.9\textwidth}{!}{
        \begin{tabular}{l|cccc|cccc|c}
            \toprule
            \multicolumn{1}{l|}{\bf Chunk}  &\multicolumn{4}{c|}{\bf Single-Doc QA} &\multicolumn{4}{c|}{\bf Multi-Doc QA} &\multirow{2}*{\bf Avg} \\
            \multicolumn{1}{l|}{\bf Method}  &NarrativeQA &Qasper* &MFQA-en   &MFQA-zh   &HotpotQA  &2WikiM    &MuSiQue    & DuReader  & \\
            \midrule
            \multicolumn{10}{c}{\bf Llama3.1-8B}  \\
            \hline
            FC200   &\bf24.59      &42.68  &52.54  &56.14      &\underline{56.81}  &46.66  &29.99  &30.51  &42.49      \\
            SC      &\bf24.59      &42.12  &52.10           &57.43      &54.34  &45.44  &30.24  &30.68  &42.12      \\
            LC      &22.93      &42.64  &\underline{52.65}              &\bf58.54      &55.85  &\underline{47.00}  &\underline{31.58}  &\underline{30.68}  &42.73\\
            \rowcolor{gray!30}
            HC200   &23.75   &\underline{43.57}      &\bf54.04       &\underline{57.51}      &56.52      &\bf48.29  &31.06  &30.65  &\bf 43.17      \\
            \rowcolor{gray!30}
            \hspace{0.2cm}+AM     &\underline{24.46}  &\bf43.85  &52.10  &56.65  &\bf57.27  &46.24  &\bf31.82  &\bf30.84  &\underline{42.90}      \\
            \toprule
            \multicolumn{10}{c}{\bf Qwen3-8B}  \\
            \hline
            FC200   &22.60          &\underline{44.47}      &\underline{53.46}          &57.26          &61.13      &48.63  &36.59  &27.43  &43.95      \\
            SC      &24.73         &43.69      &52.83          &58.66          &56.22      &46.77  &37.83  &\underline{27.59}  &43.54      \\
            LC      &\underline{24.55}  &43.41  &\bf54.58  &\bf59.60  &60.50  &\bf51.00  &37.37  &\bf27.60  &\bf44.83      \\
            \rowcolor{gray!30}
            HC200   &21.96          &42.38      &51.23  &58.47  &\bf62.84      &\underline{49.57}  &\underline{38.03}  &26.74  &43.90      \\
            \rowcolor{gray!30}
            \hspace{0.2cm}+AM     &\bf21.79       &\bf46.37      &52.81  &\underline{58.86}     &\underline{61.94}      &47.17  &\bf39.09  &27.28  &\underline{44.41}      \\
            \toprule
            \multicolumn{10}{c}{\bf Qwen3-32B}  \\
            \hline
            FC200   &26.09      &43.70      &\bf50.87   &60.44      &\bf63.61      &58.03  &39.40  &28.50  &46.33      \\
            SC      &26.19   &43.47      &49.54      &61.63      &61.37  &58.13     &40.65  &\bf29.34  &46.29      \\
            LC      &26.35      &\bf44.75      &50.21      &\bf63.01  &\underline{63.31}     &60.22  &\bf42.69  &\underline{28.91}  &\bf47.43      \\
            \rowcolor{gray!30}
            HC200   &\bf27.01      &44.44      &49.69      &\underline{62.16}      &61.85      &\underline{61.24}  &38.54  &28.54  &46.71      \\
            \rowcolor{gray!30}
            \hspace{0.2cm}+AM     &\underline{26.97}      &\underline{44.49}   &\underline{50.28}      &60.47  &61.67  &\bf62.37  &\underline{40.80}  &28.29  &\underline{46.92}      \\
            \hline
        \end{tabular}
    }
    \label{tab:LongBench_rag_eval}
\end{table}

\subsection{Evidence Recall under Different Token Budget}
\label{sec:context_size_er_exp}
\begin{figure}[htbp]
    \centering
    \includegraphics[width=0.99\textwidth]{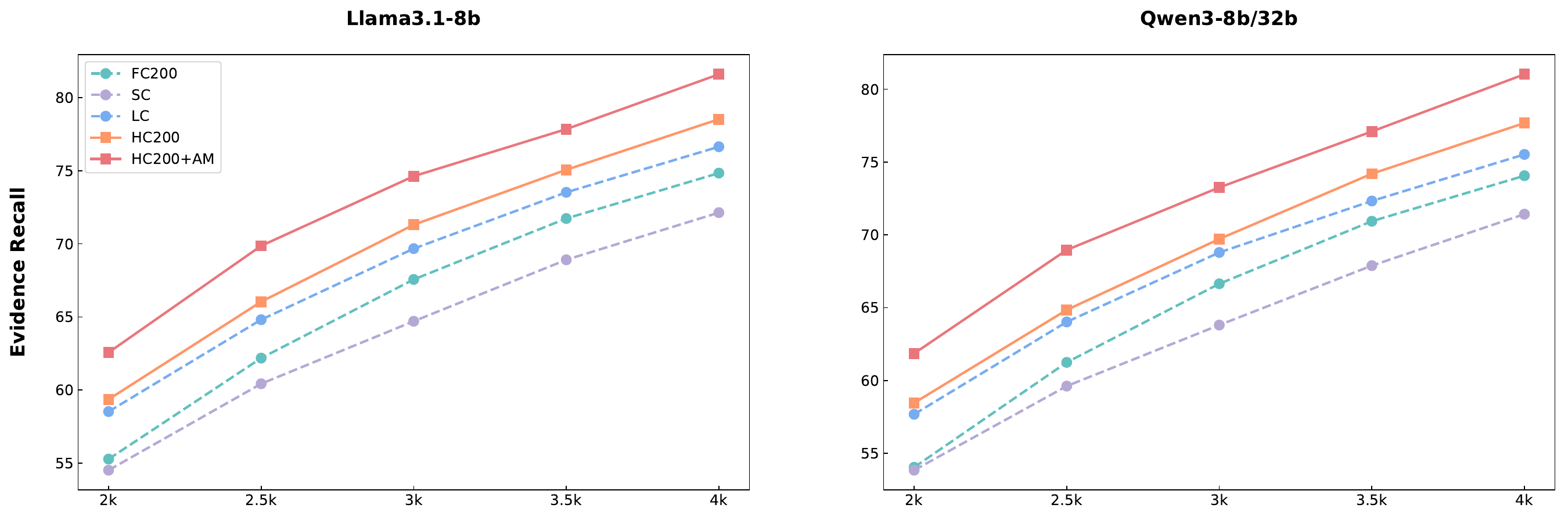}
    \caption{Evidence recall metric across different token budget on  HiCBench($T_1$).}
    \label{fig:context_size_er_exp}
\end{figure}

In this section, we further present the curve of evidence recall metric at different retrieval context length settings (from 2k to 4k). The results are shown in \autoref{fig:context_size_er_exp}. Compared with other chunking methods, the HC200+AM method always maintains the best performance.

\subsection{Prompts}
\label{app:prompts}
\begin{lstlisting}[caption=Prompt for segment summarization., label={lst:segment_summarization}]
**Task:**  
You are tasked with analyzing the provided document sections and their hierarchical structure. Your goal is to generate a concise and informative paragraph describing the content of each section and subsection.

**Instructions:**
1. Each section or subsection is identified by a header in the format `===SECTION xxx===` (for example, `===SECTION 1===`, `===SECTION 2.1===`, etc.).
2. For every section and subsection, write a brief, clear, and informative paragraph summarizing its content. Do not omit any section or subsection.
3. Present your output as a JSON object with the following structure:
```json
{
    "SECTION 1": "description of section 1",
    "SECTION 1.1": "description of section 1.1",
    ...
    "SECTION n.m": "description of section n.m"
}
```
4. Ensure that each key in the JSON object matches the exact section identifier (e.g., `"SECTION 2.1.3"`), and  do not include any sections or subsections that are not present in the provided document fragment.
5. Do not add any commentary or explanation outside the JSON object.

**Document Fragment:**
\end{lstlisting}

\begin{lstlisting}[caption=Prompt for QA cunstruction., label={lst:qa_construction}]
You are provided with a document that includes a detailed structure of sections and subsections, along with descriptions for each. Additionally, complete contents are provided for a few selected sections. Your task is to create a question and answer pair that effectively captures the essence of the selected sections. Finally, you need to extract the facts which are mentioned in the answer.

<Type of Generated Q&A Task: Evidence-dense Dependent Understanding task>
Understanding task means that, the generated question-answering pairs that require the responser to extract information from documents. The answer should be able to find directly in the documents without any reasoning.
Evidence-dense dependent means that the facts about generated question are wildly distributed across all parts of the retrieved sections.

<Criteria>
- The question MUST be detailed and be based explicitly on information in the document.
- The question MUST include at least one entity.
- Question must not contain any ambiguous references, such as 'he', 'she', 'it', 'the report', 'the paper', and 'the document'. You MUST use their complete names.
- The context sentence the question is based on MUST include the name of the entity. For example, an unacceptable context is "He won a bronze medal in the 4 * 100 m relay". An acceptable context is "Nils Sandstrom was a Swedish sprinter who competed at the 1920 Summer Olympics."
- **THE MOST IMPORTANT: Evidence-dense dependency**, Questions must require understanding of ENTIRE selected sections. Never base Q&A on isolated few sentences. For example, a question comply the **Evidence-dense dependency** criteria means that the facts about this question should be wildly distributed across all parts of the retrieved sections.

<Output Format>
Your response should be structured as follows:
```json
{{
    "question": "Your generated question here",
    "answer": "Your generated answer here"
}}
```

<Document Structure and Description>
{section_description}

<Retrieved Section and Content>
{section_content}
\end{lstlisting}

\begin{lstlisting}[caption=Prompt for evidence retrieval., label={lst:evidence_retrieval}]
**Task:**  
Analyze the relationship between context sentences and answer sentences.

**Instructions:**
1. You are given:
    - A context fragment, with each sentence numbered as follows: `[serial number]: context sentence content`
    - A question and its corresponding answer, with each answer sentence numbered as follows: `<serial number>: answer sentence content`
2. For each sentence in the answer, identify which sentence(s) from the context provide the information used to construct that answer sentence.
3. Present your findings in the following JSON format:
```json
{{
    "<answer_sentence_id_1>": "[context_sentence_id_1], ..., [context_sentence_id_n]",
    "<answer_sentence_id_2>": "[context_sentence_id_1], ..., [context_sentence_id_m]",
    ...
    "<answer_sentence_id_i>": "[context_sentence_id_1], ..., [context_sentence_id_j]"
}}
```

**Notes:**
- Only include answer sentences that have supporting evidence in the context.
- If an answer sentence does not have a source in the context, do not include it in the JSON output.
- Use only the serial numbers (not the full sentences) for both context and answer sentences in your JSON output.
- If multiple context sentences support an answer sentence, list all relevant context sentence numbers, separated by commas.

**Context Sentences:**  
{context_sentence_list}

**Question:**  
{question}

**Answer Sentences:**  
{answer_sentence_list}
\end{lstlisting}

\begin{lstlisting}[caption=Prompt for model training., label={lst:prompt_training}]
You are an assistant good at reading and formatting documents, and you are also skilled at distinguishing the semantic and logical relationships of sentences between document context. The following is a text that has already been divided into sentences. Each line is formatted as: "{line number} @ {sentence content}". You need to segment this text based on semantics and format. There are multiple levels of granularity for segmentation, the higher level number means the finer granularity of the segmentation. Please ensure that each Level One segment is semantically complete after segmentation. A Level One segment may contain multiple Level Two segments, and so on. Please incrementally output the starting line numbers of each level of segments, and determine the level of the segment, as well as whether the content of the sentence at the starting line number can be used as the title of the segment. Finally, output a list format result, where each element is in the format of: "{line number}, {segment level}, {be a title?}".

>>> Input text:
\end{lstlisting}

\end{document}